# Análisis de Canasta de mercado en supermercados mediante mapas auto-organizados

## Resumen

**Introducción:** Una cadena importante de supermercados de la zona poniente de la capital de Chile, necesita obtener información clave para tomar decisiones.
Esta información se encuentra disponible en las bases de datos, pero necesita ser procesada debido a la complejidad y cantidad de información, lo que genera una dificultad a la hora de visualizar. **Método:** Para este propósito, se ha desarrollado un algoritmo que utiliza redes neuronales artificiales, aplicando el método SOM de Kohonen. Para llevarlo a cabo, se han debido seguir ciertos procedimientos claves, como preparar la información, para luego utilizar solo los datos relevantes a las canastas de compra de la investigación. Luego de efectuado el filtrado, se tiene que preparar el ambiente de programación en Python para adaptarlo a los datos de la muestra, y luego proceder a entrenar el SOM con sus parámetros fijados luego de resultados de pruebas. **Resultado:** El resultado del SOM obtiene la relación entre los productos que más se compraron, posicionándolos topológicamente cerca, para conformar promociones y bundles, para que el retail mánager tome en consideración. **Conclusión:** En base a esto, se han hecho recomendaciones sobre canastas de compra frecuentes a la cadena de supermercados que ha proporcionado los datos utilizados en la investigación.

**Palabras clave**: SOM, Análisis de canasta de compra, Python, Minería de Datos, Redes Neuronales.

# Market basket analysis in supermarkets using self-organized maps

## *Abstract*

***Introduction:*** *An important chain of supermarkets in the western zone of the capital of Chile, needs to obtain key information to make decisions, this information is available in the databases but needs to be processed due to the complexity and quantity of information which becomes difficult to visualiz,.* ***Method:*** *For this purpose, an algorithm was developed using artificial neural networks applying Kohonen's SOM method. To carry it out, certain key procedures must be followed to develop it, such as data mining that will be responsible for filtering and then use only the relevant data for market basket analysis. After filtering the information, the data must be prepared. After data preparation, we prepared the Python programming environment to adapt it to the sample data, then proceed to train the SOM with its parameters set after test results.* ***Result:*** *the result of the SOM obtains the relationship between the products that were most purchased by positioning them topologically close, to form promotions, packs and bundles for the retail manager to take into consideration, because these relationships were obtained as a result of the SOM training with the real transactions of the clients.*
***Conclusion:*** *Based on this, recommendations on frequent shopping baskets have been made to the supermarket chain that provided the data used in the research.*

***Keywords:*** *SOM; Market Basket Analysis; Python; Data Mining; Neural Networks.*

## Introducción

El contexto de la presente investigación se refiere, al análisis del comportamiento de los clientes de supermercados al momento de comprar productos, ya que, determinados productos, habitualmente se complementan con productos predeterminados; de esta manera, analizando una gran cantidad de datos relacionados a las boletas de compra de los clientes, se puede llegar a obtener patrones de compra de productos, de ahí el nombre "Market Basket Analysis" que trata del análisis de las canastas de compra de los clientes en supermercados.
Toda esta información transaccional (como la fecha, id de transacción, productos comprados por cada cliente) representan un gran volumen de datos. Cada compra en el sistema está asociada a una transacción en particular, la cual nos ayuda a identificar los productos adquiridos dentro de esa canasta. De esta manera, podemos encontrar patrones los productos que se adquieren en conjunto. Estos patrones y relaciones pueden ser utilizados para conformar promociones y/o que eventualmente se puedan tomar decisiones con respecto a las compras de los clientes y las relaciones de los productos: se podría determinar cuáles productos en las vitrinas podrían ir acompañados con sus complementos.
Debido esta gran cantidad de información almacenada y las complejas relaciones que cada producto de esta contiene, proponemos utilizar técnicas de inteligencia artificial (en concreto, redes neuronales de tipo mapas auto-organizados) para poder analizar, trabajar, y descubrir estas relaciones.

## Problemática

En esta investigación se trató el problema de análisis de canastas de compra en una cadena de supermercado de la zona poniente de la capital de Chile, utilizando el enfoque previamente mencionado en relación con una red neuronal artificial o ANN por su sigla en inglés, demostrando que la forma de resolver el problema con este método es posible, potente y ágil, cuando se trata de una gran cantidad de datos que se quieran clasificar y determinar el comportamiento de las características que definen a las neuronas dentro de la red.

Los datos utilizados corresponden a los registros de compras de 3 meses de una cadena de supermercados ubicados en la zona poniente de Santiago la capital de Chile, la necesidad de realizar este análisis e investigación nace de encontrar promociones o bundles "no típicas" para de esta manera potenciar las ventas.

La problemática en analizar el comportamiento de compra de los clientes de una cadena de supermercados, es la cantidad de información que está asociada a las compras realizadas por los clientes y preferiblemente de un periodo no acotado de tiempo para tener una muestra aun mayor por ende más datos, la información por lo general está compuesta por miles de filas lo que complica que una persona pueda a simple vista filtrar, revisar y analizar la información en cortos periodos de tiempo y en base a esto determinar el comportamiento de compra de los clientes que se encuentra implícito en los datos. Para identificar el comportamiento entre los productos se utiliza la técnica clave de análisis de canasta de mercado (MBA) que utilizan las grandes firmas de supermercados, que consiste en descubrir las relaciones de los productos para esto busca las combinaciones de productos que se compraron frecuentemente en la misma compra para identificar el comportamiento. Una vez determinado el comportamiento de los clientes este se vuelve información clave y ventaja contra la competencia, ya que se logra saber que canastas de compra están comprando los clientes que frecuentan la cadena de supermercados que proporciono los datos, de esa forma puede saber que productos debe tener siempre en stock, ya que la ausencia de uno puede implicar en la no compra del complemento, también se puede utilizar para reducir el tiempo que los clientes pasan en el supermercado utilizando la infraestructura lo que supone en un coste para la cadena

Para darle uso a los grandes volúmenes de datos que poseen los supermercados, que se producen por cada venta que se realiza en las múltiples cajas, nace la oportunidad que consta de utilizar los datos de las ventas realizadas, para descubrir las complejas relaciones que existen entre los productos adquiridos por los clientes y esto tiene relación con que compra cada producto y en qué cantidad, para llegar a promociones en base a las relaciones encontradas.

Con los datos transaccionales del supermercado y el resultado del análisis, se determinará el comportamiento de los clientes al momento de comprar, tomando en consideración las relaciones de los productos.

## Estado del Arte

Una canasta de compra es el conjunto de productos que un cliente compra durante una sola transacción.

El análisis de canastas de mercado (i.e., market basket análysis) es un término genérico utilizado para describir las metodologías que estudian la composición de las canastas de compra adquiridas durante una sola compra. (Russell & Petersen, 2000)

### Reglas de Asociación
Una de las principales aplicaciones del análisis de canastas de compra es el método de reglas de asociación. Esta es una de las principales técnicas para detectar y extraer información útil de datos de transacciones a gran escala (Karpienko, 2016).
Existen varios trabajos en donde se hace referencia al estado del arte (Dunham, 2001) y (Bhowmick, 2003), los cuales son documentos que exponen y explican las Regla de Asociación, sus factores claves y los diferentes tipos de algoritmos existentes para llevar a cabo su procedimiento.

Las reglas de asociación pueden responder algunas preguntas tales como: qué tipo de productos tienden a ser comprados juntos por los clientes. Las reglas con una alta confianza (confidence) y un fuerte soporte (support) pueden denominarse reglas fuertes. Por ejemplo, (Chen, 2010) seleccionaron variables para la rotación de clientes de una compañía de servicio multimedia a pedido según las reglas de asociación para determinar el comportamiento de potenciales clientes y poder clasificarlos como VIP y no VIP según su nivel monetario, cantidad de veces que renueva servicios a pedido, entre otros atributos. (Beomsoo Shim, 2012) propusieron estrategias de CRM basadas en reglas de asociación y patrones secuenciales para analizar los datos de transacciones de centros comerciales en línea de pequeño tamaño.

Para obtener información desde las bases de datos, se utiliza comúnmente las reglas de asociación para buscar patrones existentes entre miles de transacciones (Agrawal, 1993).

**Mapas auto-organizados:**
Las redes neuronales artificiales (RNA o en ingles ANN), pueden ser el paradigma que más representa el aprendizaje basado en pesos. Sus fundamentos teóricos están basados en el comportamiento del sistema nervioso de los animales, es decir, un sistema de neuronas interconectado de manera colaborativa, y de esta manera puede producir un estímulo de salida o un resultado representativo.

Los SOM o mapas auto-organizados (Kohonen, 1990), son un tipo de red neuronal artificial (RNA) que utiliza el método de entrenamiento no supervisado, que se usa para reducir las dimensiones de fuentes de datos n-dimensionales, los resultados por lo general son en dos dimensiones, este mapa autoorganizado se encarga de agrupar y clusterizar los datos recibidos como entrada, para posteriormente poder clasificar cada clúster encontrado y así sus componentes.

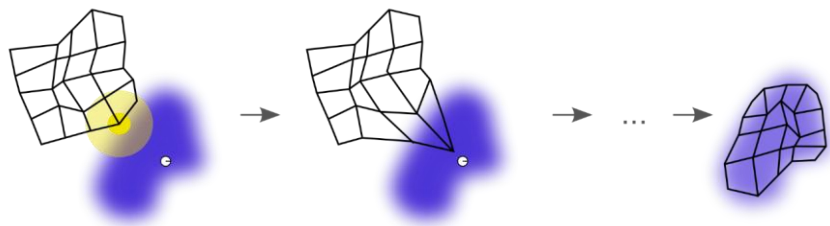

**Figura 1**. Funcionamiento de un SOM

Esta figura ilustra el funcionamiento de un SOM. La mancha azul es la distribución de los datos de entrenamiento y el pequeño disco blanco es el dato de entrenamiento actual extraído de esa distribución. Al principio (izquierda) los nodos SOM se colocan arbitrariamente en el espacio de datos. Se selecciona el nodo (resaltado en amarillo) más cercano al dato de entrenamiento extraído. Se mueve hacia el dato de entrenamiento, al igual que sus vecinos en la cuadrícula (que se mueven en menor medida). Después de muchas iteraciones, la cuadrícula tiende a aproximarse a la distribución de datos (derecha).

Los mapas auto-organizados ya han sido utilizados para analizar las canastas de compra. La investigación realizada por (Decker & Monien, 2003) reporta sobre la utilización de mapas auto-organizados para el análisis de canastas de compra en una cadena de retail alemána. Este estudio logra establecer claras relaciones de co-compra entre productos utilizando SOM. Sin embargo, creemos que el estudio tiene un contexto de aplicación muy limitado: los datos utilizados corresponden a apenas 2.000 canastas de compras de 25 productos, y estos corresponden a una tienda de artículos de higiene personal/propósito general, no a un supermercado.

En este artículo, al igual que (Decker & Monien, 2003), aplicaremos SOM a datos de canastas de compras, pero con un mayor volumen de datos y productos, un mejor análisis de sensibilidad de parámetros, y utilizando datos reales de una cadena de supermercados de la ciudad de Santiago, Chile.

## Metodología

A continuación, se presentarán los equipos y herramientas utilizadas dentro de este trabajo, para luego describir la metodología de proceso de preprocesamiento y preparación de los datos.

### Equipos

Para la realización del trabajo se utilizó un equipo portátil ASUS TUF FX504 con procesador Intel(R) Core (TM) i7-8750H CPU 2.20GHz de 6 núcleos y memoria RAM de 8GB., funcionando con Sistema operativo Windows 10 Home.
Cabe destacar que el procesamiento de datos fue ejecutado en un disco de estado sólido (SSD) para aumentar la velocidad de ejecución.

### Datos utilizados

La cadena de supermercados nos proporcionó una base de datos con compras de clientes compuesta por 146.621 filas. Cada fila hace referencia a una compra de producto individual con sus datos correspondientes.

La base de datos proporcionada contiene 9 columnas, las cuales son: Identificador de cliente, fecha de la transacción, día de la semana (1 a 7), día del mes (1 a 31) el año, categoría, subcategoría, nombre de producto y precio. La Tabla 1 muestra un extracto de datos incluidos en esta base de datos.

| ID Cliente | Fecha Transacción | Dia | Mes | Año | Categoría | Sub-Categoría | Producto | Precio |
|---|---|---|---|---|---|---|---|---|
| 429103 | 3/09/2011 | 7 | 9 | 2011 | Leche Abarrotes | Leche polvo descremada | Leche polvo descremada x 800g | 2.990 |

**Tabla 1**. Ejemplo fila Base de Datos

Se puede apreciar en la **¡Error! No se encuentra el origen de la referencia.** que el campo día no concuerda con los datos en la columna de fecha de transacción, esto a simple vista resulta extraño, pero la explicación, es que en el campo en la columna día tiene un valor numérico del día de la semana entre 1 y 7, donde 1 es lunes y 7 domingo. Además cada fila representa la venta de un producto individual, pero no hay información sobre canastas de compras: productos que se vendieron en conjunto.

Con el fin de identificar las canastas de compra en esta Base de Datos, realizamos un preprocesamiento de éstos para poder agruparlos en canastas de compra (i.e., según su fecha de transacción e ID de cliente) y además eliminar características que no serán utilizadas en el análisis.

Para realizar este preprocesamiento se utilizó la herramienta RapidMiner Studio (Team R. C., 2019), la cual permite el desarrollo de procesos de análisis y minería de datos mediante el encadenamiento de operadores los cuales ofrecen tantos parámetros configurables para cada uno, entradas y salidas. Gracias a esta útil herramienta se desarrolló un flujo de trabajo para realizar el preprocesamiento y filtrado de datos relacionados con cada compra realizada.

Este flujo es presentado en la Figura 2 y se encuentra disponible en el enlace: **https://drive.google.com/file/d/1qG0x4c4ozlIPyswDq_B_4hEyEdNBYOAL/view?usp=sharing**

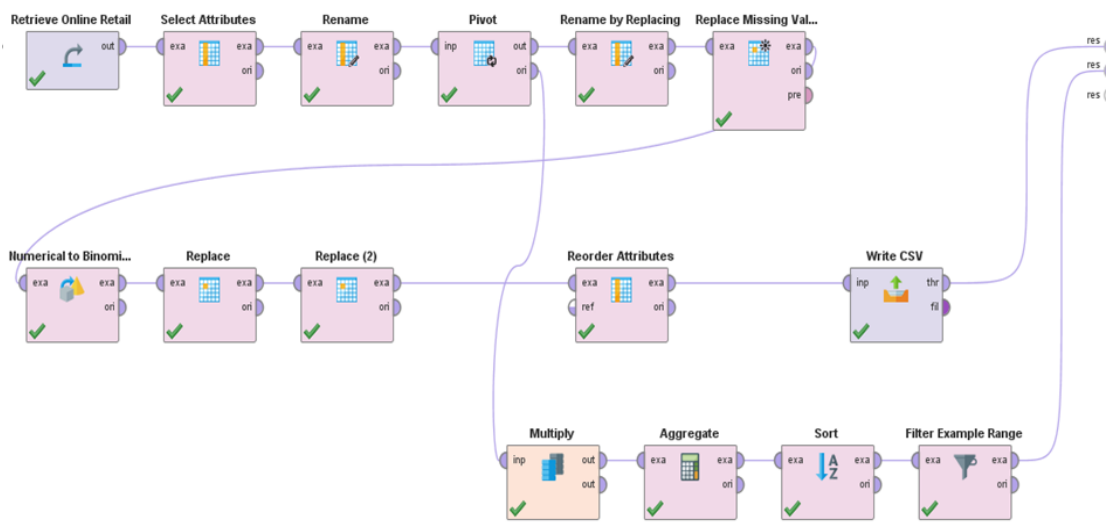

**Figura 2**. Flujo de preprocesamiento de datos en RapidMiner Studio

En este flujo asumimos que los productos comprados el mismo día por el mismo cliente corresponden a una misma canasta de compra. Posteriormente utilizamos un script en Pyton (disponible en **https://github.com/pitfox94/SOM**) para remover y renombrar columnas.
Utilizando este supuesto pudimos transformar los datos presentados en la Tabla 1 en vectores de canastas de compra como se muestra a continuación en la Tabla 2.

| Id Transacción | Fecha | Atún | Fideos | Arroz grado 2 | Yogur | Aceite |
|---|---|---|---|---|---|---|
| 001 | 07/06/2011 | 0 | 1 | 1 | 0 | 1 |

**Tabla 2** - Vector de canasta de compra

En la Tabla 2, cada fila corresponde a una transacción o "canasta de compra", en donde un valor de 0 o 1 indica la ausencia o presencia de un producto en esa canasta. Debemos notar que estos vectores de compra son largos, ya que tienen una "coordenada" por cada producto disponible en el catálogo.

Luego de pre-procesar los datos de la cadena de supermercados proporcionada en la base de datos, se observaron **48.358 transacciones o canastas de compra en 3 meses**: 17.542 transacciones de compras realizadas por los clientes en el 1° mes, 15.089 en el 2° y 15.517 en el 3° mes.

## Desarrollo

Una de las características que distinguen una red Kohonen es que las relaciones topológicas (i.e., de vecindad) de las canastas de compra utilizadas como parámetro de entrada, se reflejan en la disposición de las correspondientes celdas o unidades en la red neuronal artificial (Riaer y Schulten 1986). El objetivo del análisis es agrupar canastas "similares", es decir canastas que tienen características en común en subconjuntos disjuntos llamados clúster.

Si la propiedad que hace que la topología del mapa de Kohonen se conserve, los agrupa según el parámetro de entrada.

Es decir, los subconjuntos de datos, que son vecinos cercanos, deben asignarse en la red o grilla del SOM conservando una relación cercana. Cualquier grupo de los vectores utilizados también deberían aparecer en la red de menor dimensionalidad, en caso de que el producto sea comprado en su mayoría individualmente, como la "Cola".

Para investigar las capacidades de agrupamiento de un mapa de Kohonen (SOM), utilizamos los datos y parámetros de la siguiente manera:

El total de transacciones de la muestra está compuesto por un total de 48.359 vectores transaccionales en donde cada vector representa una canasta de compra de manera binaria donde cada vector tiene un largo de 189 productos. Se deben inicializar valores en la red o grilla del SOM con vectores binarios aleatorios con el mismo formato para poder hacer el proceso de comparación. Los números aleatorios son generados por la misma semilla gracias a la librería "numpy" para tener resultados determinísticos.

La precisión del resultado del entrenamiento del SOM en comparación con la muestra de datos reales esta por sobre el 90% es decir, que las relaciones halladas en las canastas de compras obtenidas del mapa entrenado tienen un 90% de precisión al momento de comparar cada una de las canastas obtenidas con canastas aleatoriamente seleccionadas de los datos reales.

A lo largo de la investigación se realizaron pruebas en Python antes de comenzar con los datos proporcionados por la cadena de supermercados, en un principio se desarrollaron y realizaron pruebas con los 48.359 vectores proporcionados para identificar diversos errores dentro del código y corregirlos, reportar avances inmediatos y realizar depuración (i.e., debugging) para entender la lógica y funcionamiento del algoritmo en desarrollo.

Estos datos luego se utilizaron para entrenar SOMs. Para este propósito, se utilizaron los siguientes parámetros: tamaño de la grilla (filas x columnas) tasa de aprendizaje e iteraciones (épocas)

Las celdas obtenidas del entrenamiento del SOM no son adecuadas directamente para identificar grupos de datos según los parámetros de entrada. A continuación, se describirá un método llamado matriz U que permite obtener una imagen más adecuada de la distribución vectorial.

La matriz-U contiene, por lo tanto, una aproximación geométrica de la distribución del vector en la red de Kohonen. Para obtener una visualización digital de cómo es esta distribución se propone utilizar una matriz-U de dos dimensiones, donde las celdas que contengan las distancias más pequeñas con sus vecinas serán graficadas en negro y mientras que las celdas más distantes serán graficadas en blanco, de manera que, mientras más oscuro sea el conjunto de neuronas o celdas significa que tienen características similares entre ellas. Los lugares blancos o que tienden a volverse más claros, indican una distancia mayor entre las celdas y, por consecuencia, un cambio de características con respecto a las neuronas o celdas que las rodean.

Por ejemplo, si se tiene una celda que contiene los siguientes valores de características (e.g., 2.0, 1.0, 1.5, 0.7) y las distancias euclidianas a las cuatro celdas vecinas son (e.g., 7.0,12.5,11.5, 5.0) luego la celda en la matriz U tiene un valor de 36 antes de promediar y luego 9 después de aplicar el promedio.

Un valor muy cercano a cero en una celda de la matriz-U indica que la celda está muy cerca de sus vecinos y, por lo tanto, las celdas vecinas conforman un grupo de características similares. La matriz-U implementada se grafica gracias a la librería "Matplotlib".

Luego de realizar pruebas con 48.359 vectores, una grilla de 20 filas y 20 columnas resultaban en una gran cantidad de neuronas vacías y con los mismos valores vectoriales. Por otra parte, una grilla compuesta por 8 filas y 8 columnas resultaba en una grilla con una agrupación de canastas muy generalizada, lo que dificultaba la detección de distintos grupos de canastas, por lo cual, una grilla de 10 filas y 12 columnas permitía la visualización de 3 grupos de canastas. Estos resultados pueden deducirse de las matrices U presentadas para cada configuración de tamaño de grilla presentadas en la Tabla 3.

| Nombre | Matriz-U 1 | Matriz-U 2 | Matriz-U 3 |
|---|---|---|---|
| Matriz | 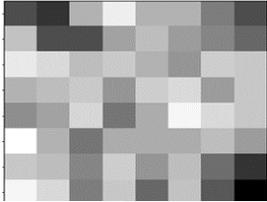 | 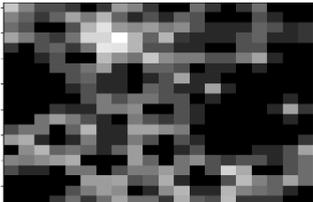 | 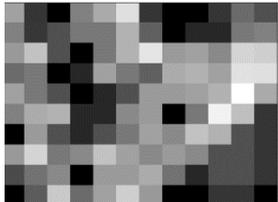 |
| Filas | 8 | 20 | 10 |
| Columnas | 8 | 20 | 12 |
| Tasa | 0,8 | 0,8 | 0,8 |
| Iteraciones | 20000 | 20000 | 20000 |

**Tabla 3**. Matrices-U para distintas dimensiones de Grilla

Las Matrices U mostradas en la Tabla 4 están conformadas por la variación de los parámetros utilizados por el programa. Las dimensiones (filas x columnas) de la grilla influyen en la posible distancia que puede tener entre cada nodo, una distancia muy grande puede generar información redundante dentro de la red y muchos clústeres poco representativos además de los clústeres 3 clústeres principales. Esto se puede apreciar en la Matriz-U 2 en la Tabla 3: una distancia pequeña genera nodos muy generalizados que tienen muchos productos dentro de sus componentes y una red en la cual resulta difícil identificar clústeres de manera gráfica.

Dado lo anterior, se optó por utilizar una dimensión de 10 x 12 en la cual se pueden apreciar 3 diferentes clústeres de manera clara.

La Tabla tiene como finalidad representar el impacto que tiene la variación de la tasa de aprendizaje en el entrenamiento de SOM, como se puede apreciar en la Matriz-U 1 y 2 una tasa muy pequeña tiene como resultados grupos que no se diferencian tanto entre si tanto como por la distancia entre los clústeres.

| Nombre | Matriz-U 4 | Matriz-U 5 | Matriz-U 3 |
|---|---|---|---|
| Matriz | 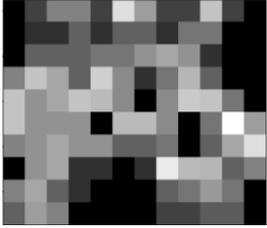 | 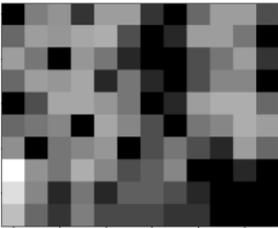 | 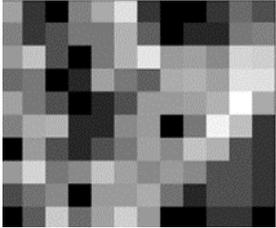 |
| Filas | 10 | 10 | 10 |
| Columnas | 12 | 12 | 12 |
| Tasa | 0,3 | 0,5 | 0,8 |
| Iteraciones | 20.000 | 20.000 | 20.000 |

**Tabla 4**. Matrices-U para distintas tasas de aprendizaje

La Matriz-U 3 de la Tabla 4 posee la ventaja de que sus clústeres encontrados son claramente visualizados gráficamente en la matriz resultante, lo que ayuda a determinar los productos que gobiernan esos clústeres y así acotar la información recibida, para que esta luego pueda ser analizada.

Para definir las dimensiones de la grilla del SOM se debe evitar perder relaciones de productos al tener una grilla muy pequeña o muchas neuronas vacías producto de una grilla muy grande. Luego de las pruebas, se determinó, implementar una red de 10 filas y 12 columnas, la tasa de aprendizaje se fijó en 0.8 y la cantidad iteraciones que, a mayor cantidad de datos, más iteraciones deben ser, la cantidad de iteraciones utilizados fue de 20.000.

Tomando en consideración lo anterior, se encontraron 3 agrupaciones o clúster importantes de productos que comparten características en común, los cuales son descritos en la Figura 3 como C1, C2 y C3. Estos grupos fueron conformados, tomando en consideración la distancia de las celdas con sus vecinas en la matriz-U.

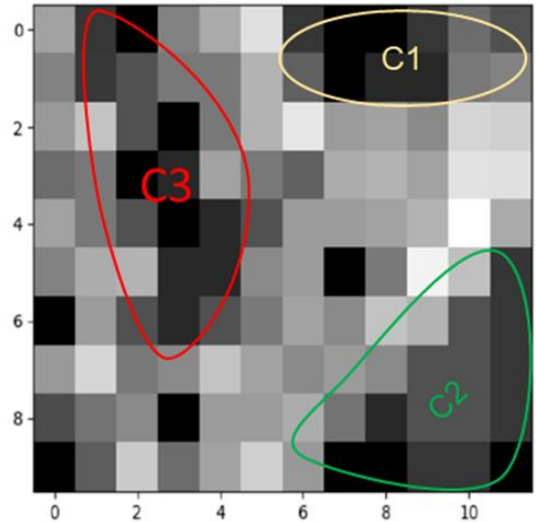

**Figura 3**. Clústeres de productos encontrados en la Matriz U 3

La Figura 4 muestra una representación alternativa, en donde en cada celda de la grilla se identifican los productos con asociaciones significativas a ella (ver Tabla 5) y al clúster al que pertenece, indicado por el color de la celda. El color fue de las celdas fue introducido por los autores para demarcar la presencia de ciertos productos emblemáticos. Notamos que sólo 22 de los 189 productos tuvieron asociaciones significativas a las celdas del SOM resultante. Esto se debe a que muchos de los 189 productos son comprados con muy poca frecuencia y luego del entrenamiento del SOM no se encontró una relación considerable.

Los valores de las celdas del mapa de la Figura 4 representan el resultado del entrenamiento en cada una de las neuronas de la grilla del SOM, el cual consiste en determinar si éstos han sido comprados en la misma canasta de compra una cantidad de veces, lo que sugiere un comportamiento según la base de datos utilizada para entrenar el SOM.

Las celdas vacías o en blanco, representan una significativa diferencia entre las características de cada una de las celdas, lo que manifiesta la correcta agrupación de vectores transaccionales similares. Las celdas coloreadas pertenecen a un mismo clúster.

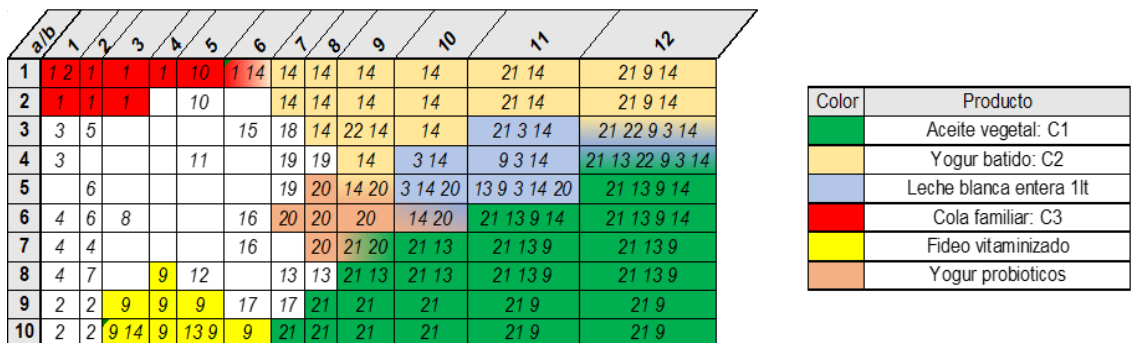

**Figura 4**. Mapa SOM 10 x 12

| ID Producto | Nombre Producto |
|---|---|
| 1 | Bebida cola familiar |
| 2 | Bebida Sabores familiar |
| 3 | Leche blanca entera natural 1 lt. |
| 4 | Trutro pollo |
| 5 | Leche sabor 1 lt |
| 6 | Vino caja tinto |
| 7 | Pack bebidas |
| 8 | Bebida Cola Light |
| 9 | Fideo vitaminizado |
| 10 | Yogur con cereales |
| 11 | Yogur Bolsa |
| 12 | Quesillo Envasado |
| 13 | Arroz grado 2 |
| 14 | Yogur batido |
| 15 | Leche Sabor |
| 16 | Yogur batido con fruta |
| 17 | Queso Crema |
| 18 | Leche polvo entera |
| 19 | Queso laminado granel |
| 20 | Yogur probiótico |
| 21 | Aceite vegetal |
| 22 | Cloro tradicional |

**Tabla 5.** Tabla Id Productos

En la Figura 4 se pueden visualizar todos los índices de productos de la Tabla 5 y además se pueden identificar los tres clústeres encontrados en la Matriz U 3 de la Figura 3, los cuales corresponden a de Aceite vegetal, Yogur batido y Cola familiar respectivamente,

## Resultados y Discusión

A continuación, analizaremos cada uno de los clústeres encontrados anteriormente.

Con respecto al clúster C1 (ver Figura 5), podemos notar que el elemento predominante en este clúster es el **aceite vegetal**, con una presencia en un 16.5% de las canastas de compra.

Podemos notar que los productos que se compran más comúnmente en conjunto con el aceite vegetal son los fideos (43%) vitaminizados y luego el arroz grado 2 (34%). Esto tiene completo sentido ya que el aceite es un producto complementario a los fideos o arroz, ya que es un insumo necesario para cocinarlos.

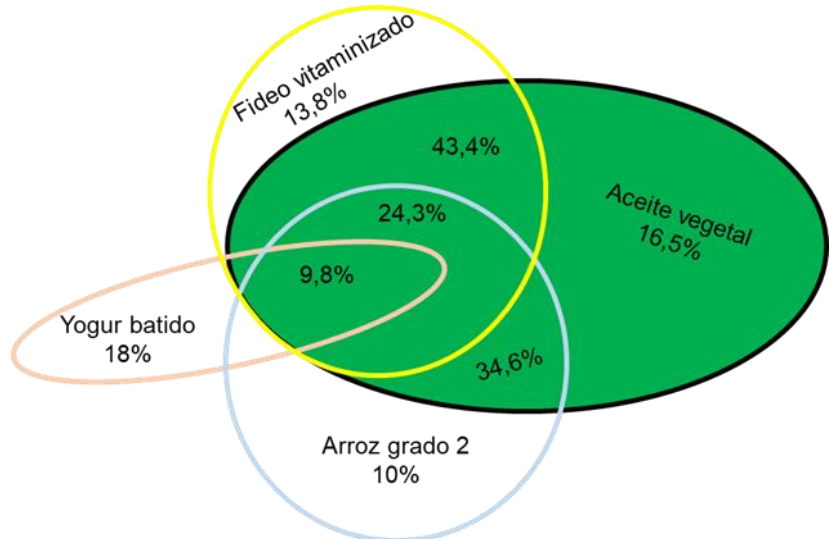

**Figura 5.** Relaciones aceite vegetal

Con respecto al clúster C2 (ver Figura 6), el elemento predominante y más comúnmente comprado en este clúster es el yogur batido, el cual está presente en un 18.2% de las canastas de compra totales.

Podemos observar que de las canastas de compra que incluyen yogur batido, casi un cuarto (24.1%) también incluye aceite vegetal. Esto puede ser explicado por el hecho de que el aceite es un producto de primera necesidad, y aparentemente aparece e gran cantidad de canastas básicas, las cuales también incluyen el yogur.

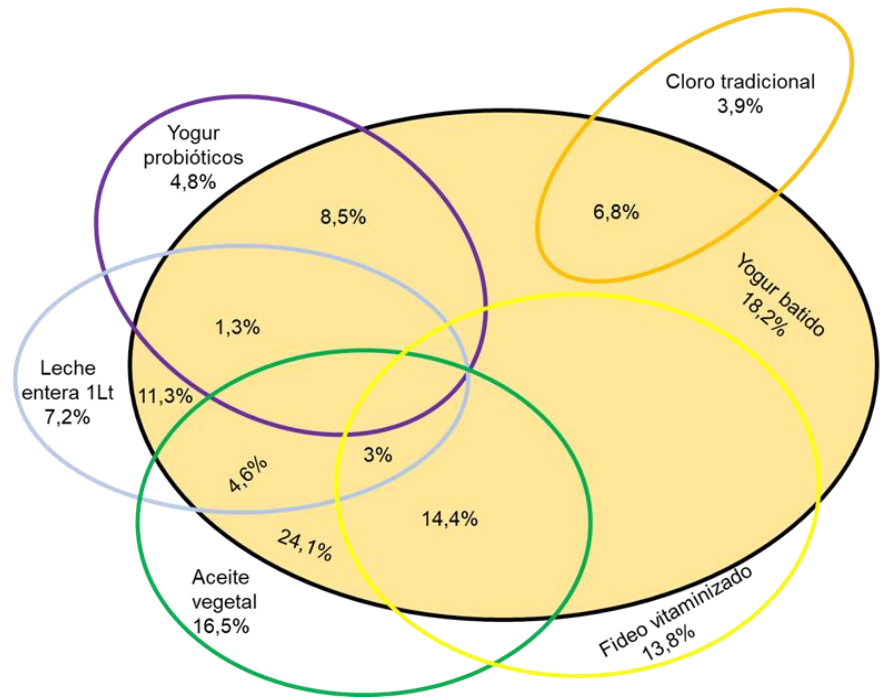

**Figura 6**. Relaciones yogur batido

Con respecto al clúster C3 (ver Figura 7), el elemento predominante y más comúnmente comprado en este clúster es la bebida cola familiar, el cual está presente en un 11.6% de las canastas de compra totales.

Las relaciones halladas para el producto "Cola familiar" y señaladas en la Figura 7 son bastante intuitivas y sencillas, ya que el hecho de que su complemento ideal sean las bebidas de sabores familiar ya se puede apreciar en promociones y packs de bebidas de cola y sabores en los supermercados actuales en Chile.

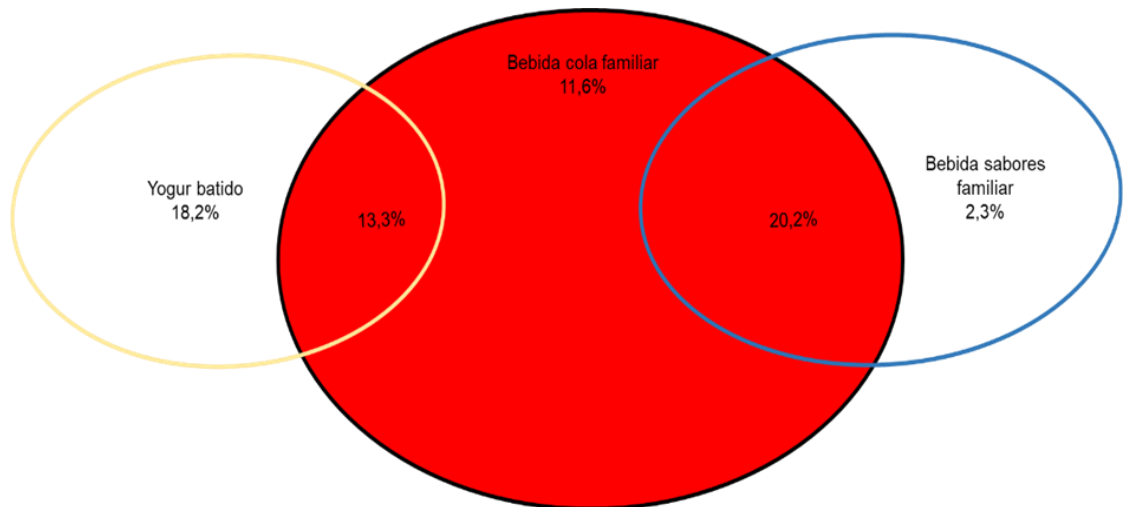

**Figura 7**. Relaciones cola familiar

## Recomendaciones

En base a los resultados obtenidos se puede realizar las siguientes recomendaciones a la cadena de supermercados:

• El aceite vegetal, con los fideos vitaminados y el arroz, deben ser colocados cada uno cerca del otro en las estanterías, ya que es altamente probable que estos productos se adquieran en una sola compra, por lo que, también se recomienda conformar packs de promociones que conformen estos 3 productos en su conjunto.

• El yogur batido, se ve complementado con el aceite vegetal y a su vez con los fideos, por lo que los convierten en una buena combinación para localizar el yogur batido cerca del aceite y fideos o de promociones de estos mismos, la relación debe estar dada por la cantidad de salsas para fideos que involucran el yogur.

• Para la bebida cola familiar, se recomienda conformar packs que conformen bebida cola familiar en el doble de la cantidad de las bebidas sabores familiares y también localizar estos productos uno cerca del otro, para la inmediata adquisición de los complementos.

## Discusión

Al ser las redes neuronales artificiales y el análisis de datos es en general un método en auge, a lo largo del tiempo se han ido desarrollando nuevas técnicas, tecnologías y librerías, para llevar a cabo de aun mejor manera para obtener aun mejores resultados, con mas confianza y facilitando la visualización con cada actualización disponible, por lo que, esta investigación podría seguir extendiéndose con el uso de estas nuevas tecnologías, ya que al igual que el análisis de datos el lenguaje de programación Python cada vez es más utilizado tanto como por compañías como por instituciones académicas, por la versatilidad de su programación para realizar diferentes tipos de tareas, tanto así que se podrían automatizar la creación de SOM's con datos obtenidos desde API's para tener un mapa en tiempo real y tomar decisiones con información fresca.

## Conclusión

Los resultados en las tablas previamente señaladas sirven para visualizar de manera didáctica, todas las interrelaciones halladas gracias al resultado del SOM.
Estas interrelaciones corresponden significativamente a los productos más comprados por los clientes en las canastas provistas por la base de datos de la cadena de supermercados.

Esta información resulta sumamente importante y útil al momento de tomar decisiones ya que, brinda una clara idea de cuales productos se están comprando juntos, debido a que las relaciones encontradas son producto de la coincidencia de compra de los clientes que frecuentan la cadena de supermercados.

La integración de variables de la "canasta" que describen el comportamiento de compra de acuerdo con el tipo de productos comprados, constituye una mejora real en el modelo predictivo de comportamiento de compra.
Sin embargo, si el número de productos es grande, esto lleva a la búsqueda de un indicador analítico que permita la caracterización de perfiles de personas estrechamente relacionadas al comportamiento, a menudo llamado "patrones".

Más allá de este uso de facilitar la comprensión global de un mercado, el comportamiento es coherente en términos de la alta frecuencia de compra de varios productos. El interés potencial por un uso predictivo se verificó en una aplicación empírica, como la del aceite vegetal con fideos y/o yogur con leche.

La ventaja de SOM por sobre otros métodos para enfrentar este tipo de problemas es que existe una alta cantidad de herramientas, aplicaciones y librerías relacionadas, por lo que su implementación resulta más eficiente que algunos otros métodos como redes neuronales de gas o NGN por sus siglas en ingles.

### Trabajo futuro
La investigación se podría seguir extendiendo debido a que se puede lograr determinar el cambio del comportamiento de la canasta de compras en diferentes fechas del año, para identificar la influencia de la temporada en la decisión de los clientes al momento de consumir determinados productos.

# Referencias